# Are Large Language Models Actually Good at Text Style Transfer?

**Sourabrata Mukherjee[1], Atul Kr. Ojha[2], Ondřej Dušek[1]**

[1]Charles University, Faculty of Mathematics and Physics, Prague, Czechia
[2]Insight SFI Research Centre for Data Analytics, DSI, University of Galway, Ireland
{mukherjee,odusek}@ufal.mff.cuni.cz
atulkumar.ojha@insight-centre.org

## Abstract

We analyze the performance of large language models (LLMs) on Text Style Transfer (TST), specifically focusing on sentiment transfer and text detoxification across three languages: English, Hindi, and Bengali. Text Style Transfer involves modifying the linguistic style of a text while preserving its core content. We evaluate the capabilities of pre-trained LLMs using zero-shot and few-shot prompting as well as parameter-efficient finetuning on publicly available datasets. Our evaluation using automatic metrics, GPT-4 and human evaluations reveals that while some prompted LLMs perform well in English, their performance in on other languages (Hindi, Bengali) remains average. However, finetuning significantly improves results compared to zero-shot and few-shot prompting, making them comparable to previous state-of-the-art. This underscores the necessity of dedicated datasets and specialized models for effective TST.

## 1 Introduction

Text style transfer (TST) involves rewriting text to incorporate additional or alternative stylistic elements while preserving its overall semantics and structure (Mukherjee and Dušek, 2024; Jin et al., 2022). Although style transfer has garnered increased research interest (Mukherjee et al., 2024a), it usually requires a substantial amount of labeled training examples, either as parallel text data (Mukherjee and Dusek, 2023) or non-parallel text data of a single style (Mukherjee et al., 2022). Recent survey papers have identified a need for new methods that reduce the training data requirements and expand the scope of styles supported (Jin et al., 2022; Hu et al., 2022b). This makes LLM prompting a compelling option and a few works explore it in TST (Liu et al., 2024a; Suzgun et al., 2022), but LLM's usefulness, particularly in multilingual and diverse stylistic contexts and with new open LLMs, requires further exploration.

This paper aims at evaluating LLMs on TST systematically. We focus on two popular subtasks of TST, sentiment transfer (Li et al., 2018) and text detoxification (Dementieva et al., 2022), and three languages: English, Hindi, and Bengali. We evaluate the LLMs using zero-shot and few-shot prompting. Additionally, we investigate parameter-efficient finetuning (Hu et al., 2022a; Mangrulkar et al., 2022). Using automatic metrics as well as human evaluation and reference-free GPT-4-based evaluation (Kocmi and Federmann, 2023), we compare our results to previous state-of-the-art (SOTA), i.e., smaller language models specifically trained on the same dedicated datasets.

Our findings indicate that GPT-3.5 as well as a few open LLMs show promising results, but do not surpass previous SOTA. While the performance of open LLMs on prompting is weaker, finetuning leads to significantly improvements, aligning closely with GPT-3.5 and SOTA performance. This highlights the necessity of dedicated datasets and models tailored for TST tasks.[1]

## 2 Related Work

TST typically involves training on pairs of texts that share content but differ in style. A standard sequence-to-sequence supervised training approach is particularly challenging due to the limited availability of parallel data (Hu et al., 2022b; Mukherjee et al., 2023a). TST methods are thus often unsupervised (Mukherjee et al., 2022; Prabhumoye et al., 2018; Li et al., 2018), which leads to high data requirements (Hu et al., 2022b).

Prompt-based methods have become popular recently, with LLM's ability to solve various downstream tasks (Brown et al., 2020; Sanh et al., 2021), including TST (Reif et al., 2021; Suzgun et al., 2022; Liu et al., 2024a). While these previous works achieved some success using non-instruction tuned models such as GPT-3, LaMDa or GPT-J, a comprehensive evaluation using different-sized instruction-tuned LLMs and prompting as well as finetuning is still needed.

---
[1]Our experimental code and other details are available at: https://github.com/souro/tst_llm.

## 3 Experiments

### 3.1 Datasets & Tasks

We use two popular TST subtasks where multilingual data is available. We selected datasets in English, Hindi, and Bengali for sentiment transfer (Mukherjee et al., 2024b, 2023a) and an English and Hindi dataset for text detoxification (Mukherjee et al., 2023b). Each dataset comprises 1,000 style-parallel examples. We use 400 examples for LLM finetuning (where applicable), 100 for development, and 500 for testing in all configurations. For sentiment transfer, experiments were conducted for both positive-to-negative and negative-to-positive tasks, with results averaged. For detoxification, we focused on the single task of transferring toxic to clean text.

### 3.2 Tested Models

For our experiments, we selected multiple freely available Language Model (LLM) architectures: BLOOM (BigScience Workshop, 2023; Muennighoff et al., 2023), ChatGLM (Du et al., 2022), Falcon (Penedo et al., 2023; Almazrouei et al., 2023), Llama (Touvron et al., 2023a,b; AI@Meta, 2024), Mistral (Jiang et al., 2023), OPT (Zhang et al., 2022), and Zephyr (Tunstall et al., 2023). They include a range of sizes (ca. ~0.5B-30B parameters) and types, including base, instruction-tuned and chat models (see Table 12 in Appendix C).[2] We also included GPT-3.5 *(gpt-3.5-turbo)* accessed via the OpenAI API (OpenAI, 2023b).[3]

For each model, we evaluate three setups: zero-shot prompts (ZS), few-shot prompts (FS), and parameter-efficient finetuning (FT). We only use the base models for finetuning, excluding chat-based and instruction-tuned models. We indicate the model variant (size, base/instructions/chat) in the model name (see Table 13 in Appendix C).[4]

As comparison to previous SOTA, we use Mukherjee et al. (2024b)'s models for sentiment transfer (*Joint* and *Parallel*) and Mukherjee et al. (2023b)'s models for text detoxification (*Seq2seq + CLS_OP* and *KT*).

---

[2]We got all models from HuggingFace (Wolf et al., 2020).
[3]As GPT-4 is used for evaluation (see Section 3.3), we did not use it for the TST task as LLMs may show bias towards their own outputs (Koo et al., 2023; Stureborg et al., 2024).
[4]More details, including prompts, are shown in Appendix B and Table 13.

| Language | Sentiment acc. (%) | Toxicity acc. (%) |
|---|---|---|
| English | 93.4 | 94.8 |
| Hindi | 89.3 | 70.9 |
| Bengali | 87.8 | - |

Table 1: Language-wise sentiment and toxicity classifier's accuracy (acc.) scores.

Due to the high cost of running LLMs, we did not conduct any extensive hyperparameter optimization. We ran limited preliminary experiments on the English and Hindi style transfer development set, opting to use default parameters from the Llama-Factory finetuning framework.[5] The only change made was increasing the number of fine-tuning epochs from 3 to 5. The same settings were then applied to both tasks and all languages.

### 3.3 Evaluation Metrics

To measure sentiment transfer and detoxification accuracy (ACC) in all experiments, we finetuned style classifiers for all languages and tasks based on *XLM-RoBERTa-base* (Conneau et al., 2020), using the training split of the same datasets. Table 1 presents the resulting classifier accuracies. In line with previous studies (Mukherjee et al., 2023c; Jin et al., 2022; Hu et al., 2022b), we evaluate content retention through the BLEU score (Papineni et al., 2002) and content similarity (CS) (Rahutomo et al., 2012) compared to the input sentences. CS is computed using LaBSE sentence embeddings (Feng et al., 2022) and cosine similarity. Following Loakman et al. (2023) and Yang and Jin (2023), we use the arithmetic mean (AVG) of ACC and CS as a singular score for comparison.

To complement automatic metrics, we employed a GPT-4-based (*gpt-4-turbo;* OpenAI, 2023a) evaluation on a sample of 50 outputs from best LLMs according to automatic metrics, following prior work that showed good correlation with humans on machine translation (Kocmi and Federmann, 2023).[6] We also conducted a small-scale in-house human evaluation on 50 outputs for best LLMs on the sentiment transfer task (for details, see Appendix D). Both humans and GPT-4 rated outputs on a 5-point Likert scale for style transfer accuracy, content preservation, and fluency.

---

[5]https://github.com/hiyouga/LLaMA-Factory
[6]Prompt details are shown in Appendix B.

## 4 Results and Analysis

### 4.1 Automatic Evaluation

We show abridged results for LLMs (with mostly ~7B variants) in Table 2. Full results are provided in Tables 6 in Appendix A.

**Impact of Methodology** GPT-3.5 consistently outperforms other models on zero-shot prompting across all languages, achieving the highest accuracy and average scores. Other models, such as ChatGLM2-6B and Llama-3-8B-ZS, also show strong performance, particularly in English. However, models like BLOOMz-7B and OPT-6.7B reach much lower scores, suggesting limited zero-shot capabilities. Few-shot prompting generally improves performance compared to zero-shot, especially in English. GPT-3.5 stays in the lead, with high scores in all languages. Finetuning brings the highest gains across the board, with strong performance from most LLMs, including ones weak at zero-shot and few-shot, such as BLOOM-7B. Most finetuned LLMs are comparable to prompted GPT-3.5 and previous SOTA models.

**Language-wise Analysis** Across the three languages, English consistently shows the highest performance. Hindi, while more challenging, benefits significantly from few-shot and finetuning approaches (e.g., for GPT-3.5 and BLOOM-7B). Bengali presents the greatest difficulty, reflecting the scarcity of high-quality training data, but still shows marked improvements with additional training. Models such as GPT-3.5 and Llama-3-8B lead in performance across all settings. The results highlight the importance of model adaptation with targeted datasets in multilingual settings.

**Impact of Model Variant** Generally, larger models score better across the board, but gains diminish with increasing size: The jump from 1B to 3B shows a significant boost; improvements from 3B to 7B and 7B to 13B are less pronounced; 30B models do not improve over their smaller counterparts. For zero-shot tasks, small models struggle, but even medium-sized models (2B-3B) show noticeable improvements. Instruction-tuned and chat models work better than their base variants in zero- and few-shot settings, but this depends on the task: for detoxification, Llama-3-8B-instruct simply refused to provide outputs.[7]

---

[7] A typical response was: "I cannot detoxify a sentence that contains sexual content. Is there something else I can help you with?"

**Style vs. Content** Different models show different sides of the tradeoff between ACC and CS, with ChatGLM2-6B and Zephyr-7B reaching high transfer accuracy but lagging on content preservation, while BLOOM-7B, Llama-3-8B-instruct or Falcon-7B are the opposite.

For additional details, see Appendix E.

### 4.2 GPT-4-based and Human Evaluation

We selected open models performing best in English for each methodology, alongside GPT-3.5 and previous SOTA, for GPT-4-based evaluation on both tasks (see Table 3). We kept the same models for human evaluation on sentiment transfer only (see Table 4). The sentiment and detoxification's output samples are shown in Table 5 and 14 (see in Appendix F) respectively.

Both evaluations show better performance for finetuned LLMs and previous SOTA, compared to prompted LLMs. In some cases, finetuned LLMs outperform GPT-3.5, particularly in terms of content preservation. Hindi and Bengali show lower performance than English, which suggests that more targeted resources for these languages are needed. This is further underscored by the fact that while English shows a decent correlation between GPT-4-based and human evaluation, this alignment is not as strong for Hindi (see Figure 1).

## 5 Conclusion

We evaluated the efficacy of LLMs for text style transfer, focusing on sentiment transfer and text detoxification across English, Hindi, and Bengali. We analyzed LLMs under zero-shot and few-shot prompting as well as with parameter-efficient finetuning. Our findings indicate that while some open LLMs exhibit promising performance in English, their multilingual capabilities are still limited. However, finetuning demonstrates significant improvements, aligning the performance of these models with previous state-of-the-art systems. Our study underscores the importance of tailored datasets and targeted models (even small-size) for this task.

In the future, we aim to expand our experiments to include more styles and languages. We will also look into alternative finetuning methods (Liu et al., 2024b; Jain et al., 2023) and advanced prompting techniques (Yao et al., 2024; Wei et al., 2022), to further improve performance.

|  | Sentiment Transfer | | | | | | | | | | | | Detoxification | | | | | | | |
|---|---|---|---|---|---|---|---|---|---|---|---|---|---|---|---|---|---|---|---|---|
|  | English | | | | Hindi | | | | Bengali | | | | English | | | | Hindi | | | |
| Models | ACC | CS | BL | AVG | ACC | CS | BL | AVG | ACC | CS | BL | AVG | ACC | CS | BL | AVG | ACC | CS | BL | AVG |
| BLOOM-7B-ZS | 37.8 | 77.4 | 39.8 | 51.6 | 26.6 | 79.4 | 39.6 | 48.6 | 34.4 | 78.8 | 30.3 | 47.8 | 8.6 | 76.1 | 39.0 | 41.2 | 52.2 | 79.1 | 39.8 | 57.0 |
| BLOOMz-7B-ZS | 26.0 | 40.3 | 12.6 | 26.3 | 31.6 | 35.9 | 4.0 | 23.9 | 35.2 | 35.1 | 2.5 | 24.2 | 14.2 | 69.1 | 34.4 | 39.2 | 64.8 | 69.8 | 30.5 | 55.0 |
| ChatGLM2-6B-ZS | 86.3 | 64.4 | 16.9 | 55.8 | 53.0 | 55.9 | 5.1 | 38.0 | 48.5 | 35.2 | 0.4 | 28.0 | 96.2 | 47.6 | 7.4 | 50.4 | 77.8 | 53.6 | 4.3 | 45.2 |
| Falcon-7B-ZS | 72.8 | 75.0 | 40.9 | 62.9 | 21.5 | 70.2 | 30.8 | 40.8 | 22.1 | 63.9 | 17.7 | 34.6 | 46.6 | 75.2 | 38.2 | 53.3 | 65.4 | 60.7 | 27.3 | 51.1 |
| GPT-3.5-ZS | 93.4 | 81.4 | 43.9 | 72.9 | 83.4 | 82.7 | 43.3 | 69.8 | 79.9 | 81.7 | 31.8 | 64.5 | 99.2 | 73.9 | 30.1 | 67.7 | 80.2 | 79.3 | 39.7 | 66.4 |
| Llama-7B-ZS | 36.8 | 65.9 | 23.3 | 42.0 | 22.2 | 80.2 | 41.4 | 47.9 | 12.0 | 78.2 | 30.9 | 40.4 | 11.6 | 73.2 | 37.0 | 40.6 | 52.6 | 79.7 | 42.4 | 58.2 |
| Llama-2-7B-ZS | 63.1 | 75.5 | 42.0 | 60.2 | 44.6 | 79.9 | 41.4 | 55.3 | 26.9 | 76.3 | 29.5 | 44.3 | 20.6 | 74.7 | 37.5 | 44.3 | 53.2 | 78.7 | 41.0 | 57.7 |
| Llama-2-Chat-7B-ZS | 94.0 | 78.0 | 38.4 | 70.1 | 65.2 | 78.5 | 37.2 | 60.3 | 39.0 | 71.6 | 21.5 | 44.0 | 82.8 | 70.4 | 25.9 | 59.7 | 61.8 | 76.9 | 38.1 | 58.9 |
| Llama-3-8B-ZS | 76.9 | 80.4 | 45.9 | 67.7 | 66.2 | 81.8 | 42.9 | 63.6 | 58.4 | 76.2 | 30.4 | 55.0 | 25.4 | 73.1 | 34.7 | 44.4 | 56.6 | 77.4 | 35.8 | 56.6 |
| Llama-3-8B-Instruct-ZS | 92.2 | 69.3 | 35.0 | 65.5 | 71.6 | 59.0 | 23.0 | 51.2 | 50.1 | 64.6 | 24.2 | 46.3 | - | - | - | - | - | - | - | - |
| Mistral-7B-Instruct-ZS | 80.8 | 65.8 | 29.3 | 58.6 | 32.2 | 78.8 | 36.4 | 49.1 | 22.8 | 74.6 | 22.6 | 40.0 | 89.4 | 72.1 | 33.1 | 64.9 | 61.8 | 72.0 | 30.8 | 54.9 |
| OPT-6.7B-ZS | 54.1 | 24.3 | 1.4 | 26.6 | 17.3 | 60.0 | 28.9 | 35.4 | 13.5 | 76.8 | 30.0 | 40.1 | 83.0 | 27.4 | 0.7 | 37.0 | 66.6 | 59.1 | 33.1 | 52.9 |
| Zephyr-7B-ZS | 85.0 | 71.4 | 23.1 | 59.8 | 66.7 | 71.6 | 31.2 | 56.5 | 55.2 | 67.5 | 20.9 | 47.9 | 96.8 | 54.6 | 13.2 | 54.9 | 71.8 | 63.7 | 21.4 | 52.3 |
| BLOOM-7B-FS | 32.1 | 78.8 | 43.5 | 51.5 | 24.5 | 80.2 | 40.1 | 48.3 | 16.9 | 77.9 | 29.6 | 41.5 | 22.4 | 77.1 | 41.1 | 46.9 | 52.0 | 79.6 | 41.6 | 57.7 |
| BLOOMz-7B-FS | 35.2 | 74.3 | 39.3 | 49.6 | 36.4 | 80.4 | 41.3 | 52.7 | 29.0 | 78.7 | 30.8 | 46.2 | 14.4 | 71.4 | 36.9 | 40.9 | 59.4 | 72.9 | 37.7 | 56.7 |
| ChatGLM2-6B-FS | 87.8 | 75.6 | 32.4 | 65.3 | 48.6 | 62.7 | 10.4 | 40.6 | 41.9 | 40.0 | 0.7 | 27.6 | 89.2 | 64.9 | 16.9 | 57.0 | 73.0 | 54.4 | 6.6 | 44.7 |
| Falcon-7B-FS | 77.6 | 79.6 | 46.2 | 67.8 | 15.9 | 78.4 | 39.8 | 44.7 | 17.8 | 73.4 | 27.3 | 39.5 | 24.2 | 75.9 | 39.9 | 46.7 | 56.4 | 75.5 | 40.2 | 57.3 |
| GPT-3.5-FS | 95.1 | 81.4 | 44.7 | 73.7 | 90.2 | 82.5 | 41.3 | 71.3 | 84.2 | 81.1 | 31.9 | 65.7 | 96.6 | 77.2 | 38.6 | 70.8 | 80.0 | 80.2 | 39.7 | 66.6 |
| Llama-7B-FS | 64.8 | 59.4 | 30.3 | 51.5 | 31.8 | 79.7 | 40.5 | 50.7 | 23.1 | 77.3 | 29.3 | 43.2 | 11.6 | 76.9 | 40.1 | 42.9 | 53.4 | 79.9 | 42.6 | 58.6 |
| Llama-2-7B-FS | 54.9 | 32.2 | 3.0 | 30.0 | 54.1 | 78.2 | 37.0 | 56.4 | 39.3 | 73.6 | 26.1 | 46.3 | 46.8 | 61.1 | 34.3 | 47.4 | 53.4 | 77.6 | 38.0 | 56.3 |
| Llama-2-Chat-7B-FS | 92.1 | 74.5 | 36.2 | 67.6 | 69.0 | 75.2 | 29.6 | 57.9 | 38.1 | 65.6 | 19.2 | 40.9 | 78.8 | 62.6 | 28.2 | 56.5 | 61.4 | 76.1 | 34.1 | 57.2 |
| Llama-3-8B-FS | 67.9 | 43.3 | 12.5 | 41.3 | 71.7 | 80.2 | 39.7 | 63.9 | 60.2 | 73.5 | 29.7 | 54.4 | 40.2 | 74.4 | 41.8 | 52.2 | 80.4 | 51.6 | 20.2 | 50.7 |
| Llama-3-8B-Instruct-FS | 52.2 | 11.1 | 1.4 | 21.6 | 1.2 | 15.7 | 0 | 5.6 | 50.0 | 14.4 | 0 | 21.5 | - | - | - | - | - | - | - | - |
| Mistral-7B-Instruct-FS | 87.3 | 77.3 | 39.7 | 68.1 | 33.7 | 77.8 | 34.2 | 48.6 | 36.5 | 75.2 | 25.4 | 45.7 | 92.2 | 74.5 | 32.6 | 66.5 | 61.2 | 76.9 | 37.4 | 58.5 |
| OPT-6.7B-FS | 33.9 | 63.4 | 28.0 | 41.8 | 11.4 | 77.5 | 39.3 | 42.7 | 15.1 | 75.8 | 29.4 | 40.1 | 11.2 | 75.4 | 39.3 | 42.0 | 57.0 | 70.6 | 37.2 | 54.9 |
| BLOOM-7B-FT | 91.2 | 80.6 | 43.2 | 71.7 | 83.9 | 81.0 | 40.4 | 68.4 | 81.7 | 75.6 | 26.3 | 61.2 | 92.4 | 75.8 | 41.7 | 70.0 | 82.0 | 76.6 | 33.8 | 64.1 |
| BLOOMz-7B-FT | 91.0 | 80.3 | 45.0 | 72.1 | 85.3 | 81.0 | 39.8 | 68.7 | 85.9 | 75.3 | 19.4 | 60.2 | 92.4 | 75.6 | 40.7 | 69.6 | 82.0 | 76.4 | 32.2 | 63.5 |
| ChatGLM2-6B-FT | 86.8 | 78.8 | 41.9 | 69.2 | 51.9 | 74.1 | 32.8 | 52.9 | 42.1 | 48.1 | 7.8 | 32.7 | 90.0 | 74.0 | 34.2 | 66.1 | 67.8 | 69.3 | 30.3 | 55.8 |
| Falcon-7B-FT | 88.3 | 79.6 | 43.1 | 70.3 | 37.7 | 76.2 | 35.8 | 49.9 | 40.2 | 51.0 | 8.3 | 33.4 | 87.6 | 73.8 | 37.8 | 66.4 | 68.8 | 61.3 | 21.4 | 50.5 |
| Llama-7B-FT | 91.5 | 81.6 | 47.2 | 73.4 | 69.4 | 78.5 | 39.4 | 62.4 | 41.9 | 76.0 | 28.4 | 48.8 | 91.8 | 76.1 | 42.4 | 70.1 | 67.4 | 73.9 | 36.2 | 59.2 |
| Llama-2-7B-FT | 92.9 | 81.2 | 46.5 | 73.5 | 77.5 | 78.6 | 39.2 | 65.1 | 56.7 | 76.1 | 27.9 | 53.6 | 92.4 | 76.2 | 43.3 | 70.6 | 68.8 | 74.6 | 36.2 | 59.9 |
| Llama-2-13B-FT | 92.0 | 82.0 | 47.3 | 73.8 | 79.6 | 80.2 | 40.0 | 66.6 | 61.2 | 77.4 | 29.4 | 56.0 | 95.6 | 76.1 | 42.8 | 71.5 | 73.8 | 75.5 | 36.3 | 61.9 |
| Llama-3-8B-FT | 92.0 | 81.4 | 46.8 | 73.4 | 85.7 | 82.1 | 42.4 | 70.1 | 81.9 | 80.2 | 32.3 | 64.8 | 96.8 | 76.9 | 45.2 | 73.0 | 83.2 | 78.0 | 37.2 | 66.1 |
| OPT-6.7B-FT | 91.7 | 80.6 | 44.5 | 72.3 | 29.1 | 76.8 | 38.3 | 48.1 | 22.5 | 76.3 | 27.6 | 42.1 | 95.8 | 76.7 | 42.2 | 71.6 | 58.2 | 76.1 | 39.8 | 58.0 |
| SOTA (*Joint*) | 84.5 | 81.5 | 46.1 | 70.7 | 78.3 | 82.5 | 43.8 | 68.2 | 80.3 | 78.0 | 28.1 | 62.1 |  |  |  |  |  |  |  |  |
| SOTA (*Parallel*) | 80.9 | 81.5 | 46.4 | 69.6 | 85.4 | 82.3 | 44.3 | 70.7 | 73.1 | 81.0 | 34.7 | 62.9 |  |  |  |  |  |  |  |  |
| SOTA (*CLS-OP*) |  |  |  |  |  |  |  |  |  |  |  |  | 91.6 | 76.6 | 44.2 | 70.8 | 65.0 | 78.2 | 39.8 | 61.0 |
| SOTA (*KT*) |  |  |  |  |  |  |  |  |  |  |  |  | 92.0 | 77.5 | 45.6 | 71.7 | 76.6 | 78.6 | 42.0 | 65.5 |

Table 2: Automatic metrics results: style accuracy (ACC), content similarity (CS), and BLEU (BL) against the source, and an average of all three (AVG). Only models close to 7B parameters in size are shown (with added GPT-3.5 and Llama-2-13B-FT, with the best sentiment transfer performance in its category), full results are in Table 6 in Appendix A. The best results in each category are highlighted in color.

|  | Sentiment transfer | | | | | | | | | Detoxification | | | | | |
|---|---|---|---|---|---|---|---|---|---|---|---|---|---|---|---|
|  | English | | | Hindi | | | Bengali | | | English | | | Hindi | | |
| Models | Sty. | Cont. | Flu. | Sty. | Cont. | Flu. | Sty. | Cont. | Flu. | Sty. | Cont. | Flu. | Sty. | Cont. | Flu. |
| GPT-3.5-ZS | 4.60 | 4.52 | 4.28 | 4.18 | 4.64 | 3.62 | 4.14 | 4.84 | 3.34 | 4.26 | 4.38 | 3.88 | 3.46 | 4.38 | 2.76 |
| Llama-2-7B-Chat-ZS | 4.96 | 4.50 | 4.26 | 3.22 | 3.74 | 2.64 | 1.50 | 2.16 | 2.20 |  |  |  |  |  |  |
| Mistral-7B-Instruct-ZS |  |  |  |  |  |  |  |  |  | 3.08 | 4.20 | 3.90 | 1.52 | 4.32 | 2.32 |
| GPT-3.5-FS | 4.68 | 4.58 | 3.92 | 4.74 | 4.60 | 3.72 | 4.42 | 4.50 | 3.22 | 4.02 | 4.72 | 3.88 | 3.44 | 4.40 | 2.94 |
| Mistral-7B-Instruct-FS | 4.16 | 4.28 | 3.98 | 2.26 | 4.00 | 3.02 | 1.78 | 3.62 | 2.62 | 3.36 | 4.66 | 3.82 | 1.62 | 3.98 | 2.18 |
| Llama-2-13B-FT | 4.70 | 4.44 | 3.96 | 4.16 | 4.20 | 3.32 | 2.98 | 3.32 | 2.60 |  |  |  |  |  |  |
| Llama-3-8B-FT |  |  |  |  |  |  |  |  |  | 3.92 | 4.44 | 3.40 | 3.22 | 4.08 | 2.88 |
| SOTA (*Joint*) | 4.14 | 4.26 | 3.56 | 4.04 | 4.60 | 3.48 | 3.62 | 4.04 | 2.84 |  |  |  |  |  |  |
| SOTA (*KT*) |  |  |  |  |  |  |  |  |  | 3.42 | 4.24 | 3.26 | 2.30 | 4.52 | 2.62 |

Table 3: GPT-4-based evaluation of 50 randomly selected outputs on style accuracy (Sty.), content preservation (Cont.), and fluency (Flu.; see Section 3.3). The best results overall are highlighted in color.

|  | English | | | Hindi | | |
|---|---|---|---|---|---|---|
| Models | Style | Content | Fluency | Style | Content | Fluency |
| GPT-3.5-ZS | 4.66 | 4.96 | 4.92 | 4.18 | 4.92 | 4.90 |
| Llama-2-7B-Chat-ZS | 4.90 | 4.86 | 4.88 | 3.54 | 4.86 | 4.22 |
| GPT-3.5-FS | 4.66 | 4.98 | 4.92 | 4.72 | 4.88 | 4.80 |
| Mistral-7B-Instruct-FS | 4.30 | 4.78 | 4.82 | 2.26 | 4.70 | 4.62 |
| Llama-2-13B-FT | 4.68 | 4.90 | 4.86 | 4.34 | 4.84 | 4.80 |
| SOTA (*Joint*) | 4.22 | 4.94 | 4.60 | 3.96 | 4.94 | 4.90 |

Table 4: Human evaluation of 50 randomly selected outputs on sentiment accuracy (Style), content preservation (Content), and Fluency (see Section 3.3). The best results overall are highlighted in color.

| Models | Negative → Positive | Positive → Negative |
|---|---|---|
| Reference | en: also matt wasn't that cool and not that helpful. → also matt wasn't that cool, and not that helpful. also matt was super cool and helpful.<br>hi: साथ ही मैट कूल-वूल नहीं था और न ही मददगार। → साथ ही मैट सुपर कूल और मददगार था।<br>bn: এছাড়াও, ম্যাট খুব একটা ভালো বা উপকারী ছিল না। → এছাড়াও, ম্যাট খুব ভালো এবং উপকারী ছিল। | en: thank you amanda, i will be back ! → no thanks amanda, i won't be back !<br>hi: धन्यवाद अमांडा, मैं वापस आऊंगा! → भाड़ में जाओ अमांडा, मैं वापस नहीं आऊंगा!<br>bn: ধন্যবাদ আমানডা, আমি আবার ফিরে আসব ! → না ধন্যবাদ আমানডা, আমি আর ফিরে আসব না ! |
| GPT-3.5-ZS | en: also matt was really cool, and very helpful.<br>hi: साथ ही मैट कूल-वूल था और मददगार।<br>bn: এছাড়াও, ম্যাট খুব একটা ভালো বা উপকারী ছিল। | en: I will not be back, Amanda, thank you.<br>hi: शुक्रिया अमांडा, मैं वापस आऊंगा!<br>bn: দুঃখিত আমানডা, আমি আবার ফিরে আসব। |
| Llama-2-7B-Chat-ZS | en: Also, Matt wasn't that cool, and he was very helpful.<br>hi: साथ ही मैट कूल-वूल था और मददगार।<br>bn: এছাড়াও, ম্যাট খুব একটা ভালো বা উপকারী ছিল না। | en: amanda, i will be back !<br>hi: निर्भया अमांडा, मैं नहीं आऊंगा!<br>bn: ধন্যবাদ আমানডা, আমি আবার ফিরে আসব ! আমানডা আমানডা, আমি আবার ফিরে আসব ! |
| GPT-3.5-FS | en: also matt was really cool and super helpful.<br>hi: साथ ही मैट कूल-वूल था और मददगार भी।<br>bn: এছাড়াও, ম্যাট খুব ভালো এবং উপকারী ছিল । | en: thank you amanda, i will never be back !<br>hi: धन्यवाद अमांडा, मैं वापस नहीं आऊंगा।<br>bn: দুঃখিত আমানডা, আমি আবার ফিরে আসব। |
| Mistral-7B-Instruct-FS | en: also matt was very cool and helpful.<br>hi: साथ ही मैट कूल-वूल था और मददगार के साथ मैट कूल-वूल हो गया।<br>bn: এছাড়াও, ম্যাট খুব একটা ভালো বা উপকারী ছিল । | en: Thank you Amanda, I won't be back.<br>hi: धन्यवाद अमांडा, मैं वापस आउंगा!<br>bn: ধন্যবাদ আমানডা, আমি আবার ফিরে আসব না । |
| Llama-2-13B-FT | en: also matt was super cool and super helpful.<br>hi: साथ ही मैट कूल-वूल था और न ही मददगार।<br>bn: এছাড়াও, ম্যাট খুব একটা ভালো বা উপকারী ছিল । | en: thanks amanda, i will never return.<br>hi: इस किसी भी व्यवसाय के लिए एक बेहतर जगह नहीं है, अमांडा!<br>bn: আমানডা, আমি আবার ফিরে আসব না ! |
| SOTA (Joint) | en: also matt was cool, and very helpful.<br>hi: साथ ही मैट बहुत साफ और मददगार था।<br>bn: এছাড়াও, ম্যাট খুব ভালো এবং উপকারী ছিল । | en: sorry amanda, i will not be back.<br>hi: धन्यवाद अमांडा, मैं वापस नहीं आऊंगा!<br>bn: ধন্যবাদ আমানডা, আমি আর ফিরে আসব না ! |

Table 5: Sample outputs for the Sentiment Transfer task (positive to negative and negative to positive) in English, Hindi, and Bengali, generated by a selection of top-performing models (see Section 4.2).

# Limitations

While our study provides insights into the performance of LLMs in TST across multiple languages, certain limitations must be considered. Our evaluation focuses on sentiment transfer and text detoxification, omitting other TST tasks, such as formality, humor, or sarcasm. Our analysis is constrained by data availability and may not fully capture the diversity of linguistic styles and cultural nuances across different languages. Finally, our study explores basic prompt techniques and finetuning, omitting advanced prompting and optimization approaches.

# Ethics Statement

In conducting this research, we adhere to ethical principles to ensure the responsible use of language models and the fair treatment of linguistic data. We prioritize transparency and accountability by documenting our methodologies, datasets, and evaluation criteria. Additionally, we respect user privacy and data confidentiality by anonymizing sensitive information and obtaining appropriate consent. Moreover, we acknowledge the potential societal impact of language models, including their potential to perpetuate biases or misinformation. Therefore, we strive to mitigate these risks by continuously evaluating and addressing ethical considerations throughout our research. Our ultimate goal is to contribute positively to advancing natural language processing while upholding ethical standards and promoting equitable access to linguistic resources and technologies.

## Acknowledgments

This research was funded by the European Union (ERC, NG-NLG, 101039303) and Charles University project SVV 260 698. We acknowledge the use of resources provided by the LINDAT/CLARIAH-CZ Research Infrastructure (Czech Ministry of Education, Youth, and Sports project No. LM2018101). We also acknowledge Panlingua Language Processing LLP for collaborating on this research project.

Atul Kr. Ojha would like to acknowledge the support of the Science Foundation Ireland (SFI) as part of Grant Number SFI/12/RC/2289_P2 Insight_2, Insight SFI Research Centre for Data Analytics.

## A  Full Experimental Results

This section presents the full set of experimental results (see Table 6), providing a detailed comparison of all methodologies and their performance across different languages and tasks. These tables offer a deeper insight into the data and support the findings discussed in the main paper. A selection of models out of it is presented in Table 2.

## B  Prompt Examples

This section provides a collection of example prompts (in English) for the Text Sentiment Transfer (Table 7) and Text Detoxification (Table 8) tasks. Additionally, we include prompts (in English) used for GPT-4-based evaluations, covering Sentiment Transfer accuracy (Tables 9), content preservation (Tables 10), and fluency (Tables 11).

|  | Sentiment Transfer | | | | | | | | | | | | Detoxification | | | | | | | |
|---|---|---|---|---|---|---|---|---|---|---|---|---|---|---|---|---|---|---|---|---|
|  | English | | | | Hindi | | | | Bengali | | | | English | | | | Hindi | | | |
| Models | ACC | CS | BL | AVG | ACC | CS | BL | AVG | ACC | CS | BL | AVG | ACC | CS | BL | AVG | ACC | CS | BL | AVG |
| BLOOM-560M-ZS | 13.3 | 65.4 | 30.1 | 36.3 | 18.9 | 64.4 | 19.9 | 34.4 | 20.8 | 60.8 | 9.3 | 30.3 | 16.6 | 70.3 | 35.2 | 40.7 | 66.0 | 63.3 | 19.5 | 49.6 |
| BLOOM-1B-ZS | 36.6 | 73.5 | 39.2 | 49.8 | 20.3 | 74.7 | 33.9 | 43.0 | 21.4 | 74.5 | 26.9 | 40.9 | 17.2 | 67.6 | 36.9 | 40.6 | 56.4 | 75.5 | 39.2 | 57.0 |
| BLOOM-3B-ZS | 44.9 | 76.7 | 41.6 | 54.4 | 36.1 | 79.4 | 40.6 | 52.1 | 40.1 | 78.5 | 30.4 | 49.7 | 10.8 | 74.9 | 40.0 | 41.9 | 52.2 | 79.5 | 42.4 | 58.0 |
| BLOOM-7B-ZS | 37.8 | 77.4 | 39.8 | 51.6 | 26.6 | 79.4 | 39.6 | 48.6 | 34.4 | 78.8 | 30.3 | 47.8 | 8.6 | 76.1 | 39.0 | 41.2 | 52.2 | 79.1 | 39.8 | 57.0 |
| BLOOMz-560M-ZS | 46.4 | 18.8 | 2.5 | 22.6 | 25.8 | 32.5 | 4.3 | 20.9 | 37.1 | 35.2 | 4.1 | 25.5 | 10.2 | 75.9 | 38.3 | 41.4 | 69.2 | 66.9 | 24.1 | 53.4 |
| BLOOMz-1B-ZS | 46.2 | 14.0 | 0.0 | 20.1 | 47.7 | 18.6 | 0.0 | 22.1 | 35.3 | 23.8 | 1.3 | 20.1 | 13.0 | 72.9 | 34.4 | 40.1 | 57.4 | 73.6 | 36.7 | 55.9 |
| BLOOMz-3B-ZS | 48.6 | 17.9 | 0.2 | 22.2 | 49.1 | 22.7 | 0.2 | 24.0 | 43.6 | 24.2 | 0.4 | 22.7 | 11.0 | 74.8 | 38.2 | 41.3 | 54.4 | 77.7 | 41.2 | 57.8 |
| BLOOMz-7B-ZS | 26.0 | 40.3 | 12.6 | 26.3 | 31.6 | 35.9 | 4.0 | 23.9 | 35.2 | 35.1 | 2.5 | 24.2 | 14.2 | 69.1 | 34.4 | 39.2 | 64.8 | 69.8 | 30.5 | 55.0 |
| ChatGLM-6B-ZS | 84.9 | 69.8 | 25.5 | 60.1 | 40.6 | 39.0 | 1.6 | 27.1 | 38.6 | 35.1 | 1.3 | 25.0 | 89.4 | 59.2 | 11.0 | 53.2 | 83.2 | 25.2 | 0.6 | 36.3 |
| ChatGLM2-6B-ZS | 86.3 | 64.4 | 16.9 | 55.8 | 53.0 | 55.9 | 5.1 | 38.0 | 48.5 | 35.2 | 0.4 | 28.0 | 96.2 | 47.6 | 7.4 | 50.4 | 77.8 | 53.6 | 4.3 | 45.2 |
| Falcon-7B-ZS | 72.8 | 75.0 | 40.9 | 62.9 | 22.1 | 70.2 | 30.8 | 40.8 | 22.1 | 63.9 | 17.7 | 34.6 | 46.6 | 75.2 | 38.2 | 53.3 | 65.4 | 60.7 | 27.3 | 51.1 |
| GPT-3.5-ZS | 93.4 | 81.4 | 43.9 | 72.9 | 83.4 | 82.7 | 43.3 | 69.8 | 79.9 | 81.7 | 31.8 | 64.5 | 99.2 | 73.9 | 30.1 | 67.7 | 80.2 | 79.3 | 39.7 | 66.4 |
| Llama-7B-ZS | 36.6 | 65.9 | 23.3 | 42.0 | 22.2 | 80.2 | 41.4 | 47.9 | 12.0 | 78.2 | 30.9 | 40.4 | 11.6 | 73.2 | 37.0 | 40.6 | 52.6 | 79.7 | 42.4 | 58.2 |
| Llama-13B-ZS | 57.8 | 76.7 | 43.4 | 59.3 | 54.3 | 81.0 | 41.8 | 59.0 | 25.9 | 78.6 | 30.7 | 45.0 | 22.8 | 70.1 | 36.8 | 43.2 | 52.6 | 79.9 | 42.5 | 58.3 |
| Llama-30B-ZS | 82.9 | 75.5 | 44.8 | 67.7 | 60.0 | 81.8 | 43.2 | 61.7 | 35.9 | 77.9 | 30.3 | 48.1 | 21.8 | 73.8 | 39.9 | 45.1 | 53.0 | 79.6 | 42.3 | 58.3 |
| Llama-2-7B-ZS | 63.1 | 75.5 | 42.0 | 60.2 | 44.6 | 79.9 | 41.4 | 55.3 | 26.9 | 76.6 | 29.5 | 44.3 | 20.6 | 74.7 | 37.5 | 44.3 | 53.2 | 78.7 | 41.0 | 57.7 |
| Llama-2-13B-ZS | 69.7 | 77.4 | 45.2 | 64.1 | 57.9 | 81.1 | 42.3 | 60.4 | 32.2 | 78.0 | 30.1 | 46.8 | 19.4 | 74.3 | 40.3 | 44.7 | 54.0 | 78.9 | 41.6 | 58.2 |
| Llama-2-Chat-7B-ZS | 94.0 | 78.0 | 38.4 | 70.1 | 65.2 | 78.5 | 37.2 | 60.3 | 39.0 | 71.6 | 21.5 | 44.0 | 82.8 | 70.4 | 25.9 | 59.7 | 61.8 | 76.9 | 38.1 | 58.9 |
| Llama-2-Chat-13B-ZS | 92.2 | 77.2 | 39.6 | 69.7 | 75.1 | 78.9 | 35.2 | 63.1 | 42.3 | 73.8 | 24.7 | 46.9 | 90.0 | 54.1 | 24.0 | 56.0 | 60.8 | 77.1 | 36.7 | 58.2 |
| Llama-3-8B-ZS | 76.9 | 80.4 | 45.9 | 67.7 | 66.2 | 81.8 | 42.9 | 63.6 | 58.4 | 76.2 | 30.4 | 55.0 | 25.4 | 73.1 | 34.7 | 44.4 | 56.6 | 77.4 | 35.8 | 56.6 |
| Llama-3-8B-Instruct-ZS | 92.2 | 69.3 | 35.0 | 65.5 | 71.6 | 59.0 | 23.0 | 51.2 | 50.1 | 64.6 | 24.2 | 46.3 | - | - | - | - | - | - | - | - |
| Mistral-7B-Instruct-ZS | 80.8 | 65.8 | 29.3 | 58.6 | 32.2 | 78.8 | 36.4 | 49.1 | 22.8 | 74.6 | 22.6 | 40.0 | 89.4 | 72.1 | 33.1 | 64.9 | 61.8 | 72.0 | 30.8 | 54.9 |
| OPT-1.3B-ZS | 43.7 | 26.1 | 0.3 | 23.4 | 16.9 | 63.5 | 31.7 | 37.4 | 14.5 | 76.0 | 28.9 | 39.8 | 96.6 | 19.3 | 0.0 | 38.6 | 59.4 | 68.8 | 37.5 | 55.3 |
| OPT-2.3B-ZS | 47.2 | 25.3 | 0.2 | 24.2 | 14.7 | 66.6 | 30.2 | 37.2 | 14.7 | 75.1 | 28.9 | 39.6 | 91.2 | 23.6 | 0.0 | 38.3 | 60.2 | 66.2 | 32.6 | 53.2 |
| OPT-6.7B-ZS | 54.1 | 24.3 | 1.4 | 26.6 | 17.3 | 60.0 | 28.9 | 35.4 | 13.5 | 76.8 | 30.0 | 40.1 | 83.0 | 27.4 | 0.7 | 37.0 | 66.6 | 59.1 | 33.1 | 52.9 |
| OPT-13B-ZS | 48.3 | 61.9 | 30.4 | 46.8 | 11.2 | 78.1 | 40.1 | 43.1 | 13.8 | 77.0 | 30.3 | 40.4 | 55.4 | 52.3 | 27.4 | 45.1 | 55.2 | 76.2 | 40.1 | 57.2 |
| OPT-30B-ZS | 64.1 | 45.1 | 17.8 | 42.3 | 11.1 | 71.7 | 34.5 | 39.1 | 14.1 | 76.1 | 29.8 | 40.0 | 21.2 | 69.3 | 40.6 | 43.7 | 92.8 | 12.4 | 3.7 | 36.3 |
| Zephyr-7B-ZS | 85.0 | 71.4 | 23.1 | 59.8 | 66.7 | 71.6 | 31.2 | 56.5 | 55.2 | 67.5 | 20.9 | 47.9 | 96.8 | 54.6 | 13.2 | 54.9 | 71.8 | 63.7 | 21.4 | 52.3 |
| BLOOM-560M-FS | 7.5 | 76.2 | 40.7 | 41.5 | 11.6 | 78.4 | 39.0 | 43.0 | 13.1 | 77.6 | 29.7 | 40.1 | 35.0 | 75.8 | 41.5 | 50.7 | 55.6 | 77.1 | 40.3 | 57.7 |
| BLOOM-1B-FS | 13.4 | 77.5 | 41.5 | 44.1 | 13.3 | 79.2 | 39.7 | 44.1 | 13.4 | 77.8 | 29.5 | 40.2 | 9.8 | 76.2 | 40.1 | 42.0 | 54.8 | 78.6 | 40.6 | 58.0 |
| BLOOM-3B-FS | 38.2 | 78.4 | 42.8 | 53.1 | 32.9 | 80.0 | 40.4 | 51.1 | 33.7 | 78.7 | 30.3 | 47.5 | 31.0 | 76.9 | 41.5 | 49.8 | 52.2 | 79.3 | 37.5 | 56.3 |
| BLOOM-7B-FS | 32.1 | 78.8 | 43.5 | 51.5 | 24.5 | 80.2 | 40.1 | 48.3 | 16.9 | 77.9 | 29.6 | 41.5 | 22.4 | 77.1 | 41.1 | 46.9 | 52.0 | 79.6 | 41.6 | 57.7 |
| BLOOMz-560M-FS | 39.9 | 24.4 | 4.8 | 23.0 | 20.2 | 66.2 | 24.3 | 36.9 | 20.5 | 65.5 | 18.9 | 35.0 | 14.8 | 72.2 | 37.0 | 41.3 | 53.2 | 76.8 | 38.0 | 56.0 |
| BLOOMz-1B-FS | 33.6 | 65.9 | 36.7 | 45.4 | 13.8 | 79.2 | 40.7 | 44.6 | 18.3 | 77.3 | 28.7 | 41.4 | 33.6 | 70.0 | 29.3 | 44.3 | 53.4 | 76.5 | 35.4 | 55.1 |
| BLOOMz-3B-FS | 44.1 | 53.6 | 23.5 | 40.4 | 29.9 | 74.7 | 36.4 | 47.0 | 21.9 | 75.5 | 28.2 | 41.9 | 17.2 | 73.2 | 39.4 | 43.3 | 52.0 | 77.9 | 40.0 | 56.6 |
| BLOOMz-7B-FS | 35.2 | 74.3 | 39.3 | 49.6 | 36.4 | 80.4 | 41.3 | 52.7 | 29.0 | 77.8 | 30.8 | 46.2 | 14.4 | 71.4 | 36.9 | 40.9 | 59.4 | 72.9 | 37.7 | 56.7 |
| ChatGLM-6B-FS | 81.0 | 71.5 | 28.2 | 60.2 | 36.2 | 41.8 | 2.2 | 26.7 | 41.6 | 32.3 | 1.8 | 25.2 | 89.2 | 65.6 | 16.3 | 57.0 | 81.0 | 23.5 | 0.4 | 35.0 |
| ChatGLM2-6B-FS | 87.8 | 75.6 | 32.4 | 65.3 | 48.6 | 62.7 | 10.4 | 40.6 | 41.9 | 40.0 | 0.7 | 27.6 | 89.2 | 64.9 | 16.9 | 57.0 | 73.0 | 54.4 | 6.6 | 44.7 |
| Falcon-7B-FS | 77.6 | 79.6 | 46.2 | 67.8 | 15.9 | 78.4 | 39.8 | 44.7 | 17.8 | 73.4 | 27.3 | 39.5 | 24.2 | 75.9 | 39.9 | 46.7 | 56.4 | 75.5 | 40.2 | 57.3 |
| GPT-3.5-FS | 95.1 | 81.4 | 44.7 | 73.7 | 90.2 | 82.5 | 41.3 | 71.3 | 84.2 | 81.1 | 31.9 | 65.7 | 96.6 | 77.2 | 38.6 | 70.8 | 80.0 | 80.2 | 39.7 | 66.6 |
| Llama-7B-FS | 64.8 | 59.4 | 30.3 | 51.5 | 31.8 | 79.7 | 40.5 | 50.7 | 23.1 | 77.3 | 29.3 | 43.2 | 11.6 | 76.9 | 40.1 | 42.9 | 53.4 | 79.9 | 42.6 | 58.6 |
| Llama-13B-FS | 75.4 | 77.2 | 45.8 | 66.1 | 45.9 | 80.0 | 39.6 | 55.2 | 33.9 | 77.0 | 29.2 | 46.7 | 10.4 | 77.0 | 40.4 | 42.6 | 51.6 | 79.1 | 40.3 | 57.0 |
| Llama-30B-FS | 51.3 | 19.8 | 0.0 | 23.7 | 50.2 | 81.5 | 42.3 | 58.0 | 22.6 | 77.7 | 30.8 | 43.7 | 21.0 | 73.9 | 41.2 | 45.4 | 56.4 | 78.6 | 41.5 | 58.9 |
| Llama-2-7B-FS | 54.9 | 32.2 | 3.0 | 30.0 | 54.1 | 78.2 | 37.0 | 56.4 | 39.3 | 73.6 | 26.1 | 46.3 | 46.8 | 61.1 | 34.3 | 47.4 | 53.4 | 77.6 | 38.0 | 56.3 |
| Llama-2-13B-FS | 52.7 | 24.8 | 0.1 | 25.8 | 49.4 | 78.4 | 37.4 | 55.1 | 35.6 | 76.0 | 29.0 | 46.9 | 82.8 | 31.6 | 3.4 | 39.3 | 55.0 | 78.5 | 38.6 | 57.4 |
| Llama-2-Chat-7B-FS | 92.1 | 74.5 | 36.2 | 67.6 | 69.0 | 75.2 | 29.6 | 57.9 | 38.1 | 65.6 | 19.2 | 40.9 | 78.8 | 62.6 | 28.2 | 56.5 | 61.4 | 76.1 | 34.1 | 57.2 |
| Llama-2-Chat-13B-FS | 88.0 | 65.7 | 15.7 | 56.5 | 77.2 | 75.6 | 29.6 | 60.8 | 46.6 | 71.2 | 23.0 | 46.9 | 83.4 | 55.1 | 17.2 | 51.9 | 68.4 | 76.4 | 33.1 | 59.3 |
| Llama-3-8B-FS | 67.9 | 43.3 | 12.5 | 41.3 | 71.7 | 80.2 | 39.7 | 63.9 | 60.2 | 73.5 | 29.7 | 54.4 | 40.2 | 74.4 | 41.8 | 52.2 | 80.4 | 51.6 | 20.2 | 50.7 |
| Llama-3-8B-Instruct-FS | 52.2 | 11.1 | 1.4 | 21.6 | 1.2 | 15.7 | 0 | 5.6 | 50.0 | 14.4 | 0 | 21.5 | - | - | - | - | - | - | - | - |
| Mistral-7B-Instruct-FS | 87.3 | 77.3 | 39.7 | 68.1 | 33.7 | 78.8 | 34.2 | 48.6 | 36.5 | 75.2 | 25.4 | 45.7 | 92.2 | 74.5 | 32.6 | 66.5 | 61.2 | 76.9 | 37.4 | 58.5 |
| OPT-1.3B-FS | 24.1 | 54.0 | 21.6 | 33.2 | 12.0 | 78.5 | 39.4 | 43.3 | 14.5 | 74.6 | 27.6 | 38.9 | 41.0 | 53.8 | 30.8 | 41.9 | 56.2 | 72.8 | 36.9 | 55.3 |
| OPT-2.3B-FS | 41.8 | 51.9 | 20.5 | 38.1 | 20.9 | 54.7 | 29.6 | 35.1 | 14.3 | 75.1 | 28.5 | 39.3 | 21.6 | 67.2 | 38.3 | 42.4 | 58.8 | 65.8 | 30.1 | 51.6 |
| OPT-6.7B-FS | 33.9 | 63.4 | 28.0 | 41.8 | 11.4 | 77.5 | 39.3 | 42.7 | 15.1 | 75.8 | 29.4 | 40.1 | 11.2 | 75.4 | 39.3 | 42.0 | 57.0 | 70.6 | 37.2 | 54.9 |
| BLOOM-560M-FT | 84.2 | 75.8 | 35.9 | 65.3 | 70.9 | 76.5 | 33.9 | 60.4 | 70.5 | 68.0 | 14.6 | 51.0 | 88.2 | 71.2 | 34.9 | 64.8 | 72.8 | 69.4 | 29.7 | 57.3 |
| BLOOM-1B-FT | 87.7 | 79.0 | 42.7 | 69.8 | 79.3 | 80.2 | 35.2 | 64.9 | 80.3 | 75.8 | 22.8 | 59.6 | 89.4 | 74.2 | 38.6 | 67.4 | 72.4 | 75.1 | 32.3 | 59.9 |
| BLOOMz-3B-FT | 90.0 | 80.0 | 44.0 | 71.4 | 78.9 | 80.4 | 38.0 | 65.8 | 76.3 | 77.5 | 27.2 | 60.3 | 88.2 | 75.6 | 40.6 | 68.1 | 78.6 | 75.9 | 32.9 | 62.5 |
| BLOOM-7B-FT | 91.2 | 80.6 | 43.2 | 71.7 | 83.9 | 81.0 | 40.4 | 68.4 | 81.7 | 76.5 | 26.3 | 61.2 | 92.4 | 75.8 | 41.7 | 70.0 | 82.0 | 76.6 | 33.8 | 64.1 |
| BLOOMz-560M-FT | 85.6 | 76.1 | 36.2 | 66.0 | 70.2 | 77.4 | 34.7 | 60.8 | 72.5 | 69.8 | 15.1 | 52.5 | 89.0 | 71.2 | 35.9 | 65.4 | 74.0 | 71.2 | 28.8 | 58.0 |
| BLOOMz-1B-FT | 85.8 | 79.2 | 42.4 | 69.1 | 76.2 | 80.0 | 37.3 | 64.5 | 83.7 | 74.2 | 21.3 | 59.9 | 89.0 | 74.5 | 39.7 | 67.7 | 72.2 | 74.3 | 31.1 | 59.2 |
| BLOOMz-3B-FT | 88.7 | 79.7 | 43.5 | 70.6 | 81.8 | 80.2 | 38.9 | 67.0 | 85.0 | 74.5 | 19.2 | 59.6 | 87.6 | 75.0 | 39.0 | 67.2 | 76.6 | 75.3 | 30.4 | 60.7 |
| BLOOMz-7B-FT | 91.0 | 80.3 | 45.0 | 72.1 | 85.3 | 81.0 | 39.8 | 68.7 | 85.9 | 75.3 | 19.4 | 60.2 | 92.4 | 75.6 | 40.7 | 69.6 | 82.0 | 76.4 | 32.2 | 63.5 |
| ChatGLM2-6B-FT | 86.8 | 78.8 | 41.9 | 69.2 | 51.9 | 74.1 | 32.8 | 52.9 | 42.1 | 48.1 | 7.8 | 32.7 | 90.0 | 74.0 | 34.2 | 66.1 | 67.8 | 69.3 | 30.3 | 55.8 |
| Falcon-7B-FT | 88.3 | 79.6 | 43.1 | 70.3 | 37.7 | 76.2 | 35.8 | 49.9 | 40.8 | 51.0 | 8.3 | 33.4 | 87.6 | 73.8 | 37.8 | 66.4 | 68.8 | 61.3 | 21.4 | 50.5 |
| Llama-7B-FT | 91.5 | 81.6 | 47.2 | 73.4 | 69.4 | 78.5 | 39.4 | 62.4 | 41.9 | 76.0 | 28.4 | 48.8 | 91.8 | 76.1 | 42.4 | 70.1 | 67.4 | 73.9 | 36.2 | 59.2 |
| Llama-13B-FT | 93.1 | 81.4 | 46.3 | 73.6 | 72.4 | 79.7 | 39.7 | 63.9 | 53.9 | 75.9 | 27.7 | 52.5 | 93.8 | 76.6 | 42.4 | 71.0 | 69.0 | 75.2 | 36.7 | 60.3 |
| Llama-2-7B-FT | 92.9 | 81.2 | 46.5 | 73.5 | 77.5 | 78.6 | 39.3 | 65.1 | 56.7 | 77.3 | 27.9 | 53.6 | 92.4 | 76.2 | 42.3 | 70.6 | 68.8 | 74.6 | 36.2 | 59.9 |
| Llama-2-13B-FT | 92.0 | 82.0 | 47.3 | 73.8 | 79.6 | 80.2 | 40.0 | 66.6 | 61.2 | 77.4 | 29.4 | 56.0 | 95.6 | 76.1 | 42.8 | 71.5 | 73.8 | 75.5 | 36.3 | 61.9 |
| Llama-3-8B-FT | 92.0 | 81.4 | 46.8 | 73.4 | 85.7 | 82.1 | 42.4 | 70.1 | 81.9 | 80.2 | 32.3 | 64.8 | 96.8 | 76.9 | 45.2 | 73.0 | 83.2 | 78.0 | 37.2 | 66.1 |
| OPT-1.3B-FT | 87.6 | 79.9 | 44.2 | 70.5 | 17.8 | 77.8 | 37.9 | 44.5 | 21.0 | 74.2 | 26.6 | 40.6 | 87.6 | 75.4 | 40.4 | 67.8 | 55.4 | 76.6 | 40.2 | 57.4 |
| OPT-2.7B-FT | 89.7 | 80.0 | 44.2 | 71.3 | 22.9 | 77.6 | 38.3 | 46.2 | 17.5 | 76.1 | 26.9 | 40.2 | 90.6 | 76.0 | 40.7 | 69.1 | 56.0 | 75.8 | 38.4 | 56.8 |
| OPT-6.7B-FT | 91.7 | 80.6 | 44.5 | 72.3 | 29.1 | 76.8 | 38.3 | 48.1 | 22.5 | 77.0 | 27.6 | 42.1 | 95.8 | 76.7 | 42.2 | 71.6 | 58.2 | 76.1 | 39.8 | 58.0 |
| OPT-13B-FT | 93.3 | 81.2 | 45.3 | 73.3 | 41.3 | 78.0 | 38.2 | 52.1 | 24.8 | 77.0 | 29.6 | 43.8 | 96.8 | 75.9 | 42.6 | 71.8 | 59.0 | 76.3 | 40.2 | 58.5 |
| SOTA (*Joint*) | 84.5 | 81.5 | 46.1 | 70.7 | 78.3 | 82.5 | 43.8 | 68.2 | 80.3 | 78.0 | 28.1 | 62.1 |  |  |  |  |  |  |  |  |
| SOTA (*Parallel*) | 80.9 | 81.5 | 46.4 | 69.6 | 85.4 | 82.3 | 44.3 | 70.7 | 73.1 | 81.0 | 34.7 | 62.9 |  |  |  |  |  |  |  |  |
| SOTA (*CLS-OP*) |  |  |  |  |  |  |  |  |  |  |  |  | 91.6 | 76.6 | 44.2 | 70.8 | 65.0 | 78.2 | 39.8 | 61.0 |
| SOTA (*KT*) |  |  |  |  |  |  |  |  |  |  |  |  | 92.0 | 77.5 | 45.6 | 71.7 | 76.6 | 78.6 | 42.0 | 65.7 |

Table 6: Automatic metrics results (all the models together for both the tasks): style accuracy (ACC), content similarity (CS), and BLEU (BL) against the source, and an average of all three (AVG).

| **Prompt** | Sentiment transfer changes the sentiment of a sentence while keeping non-sentiment-related content unchanged. |
|---|---|
| | Examples: |
| | Task: positive to negative<br>Input: even when she didn't answer him quickly enough, he patiently waited on her.<br>Output: when she didn't answer him quickly enough, he hung up on her. |
| | Task: negative to positive<br>Input: there isn't a great selection of salads or pasta in the deli.<br>Output: great selection of salads and pasta items in the deli. |
| | Task: positive to negative<br>Input: they were very fair and respected me since i'm so young.<br>Output: they tried to take advantage of me because i am young. |
| | Task: negative to positive<br>Input: also the crab wontons are awful along with the bland plum sauce.<br>Output: also the crab wontons rule along with the spicy plum sauce. |
| | Now change the sentiment of the following English sentence.<br>Task: positive to negative<br>Input: best customer service i have ever had.<br>Output: |

Table 7: A few-shot prompt used for Sentiment Transfer in English. It contains task definition, examples, instruction, and input (see Section 3.2).

| **Prompt** | Text detoxification changes a toxic sentence to a clean sentence while keeping its general meaning unchanged. |
|---|---|
| | Examples: |
| | Input: oh , i m hearing shit like that from her too .<br>Output: i've heard same things from her also |
| | Input: most likely he gave her a couple of pats on the butt and shes over dramatizing the whole thing .<br>Output: Most like he gave her a couple of pats on her person and she's over dramatizing the whole thing |
| | Input: no . almost nobody is saying that except for those fucking idiots in the video .<br>Output: no,almost nobody is saying that except for those in th video |
| | Input: that said , i hope they get a great attorney and sue the shit out of the cops .<br>Output: that said , i hope they get a great attorney and sue the cops . |
| | Now detoxify the following English sentence.<br>Input: DIGIT year olds can be little shits too , doesn t mean you fight them .<br>Output: |

Table 8: A few-shot prompt used for Detoxification in English. It contains task definition, examples, instruction, and input (see Section 3.2).

| **Prompt** | Sentiment transfer task: transfer the sentiment of a sentence (from positive to negative or negative to positive) while keeping the rest of the sentiment-independent content unchanged. |
|---|---|
| | Please rate the sentiment transfer accuracy of the negative to positive sentiment transfer task between the following English source sentence S1 and the sentiment-transferred sentence S2. Use a scale of 1 to 5, where 1 indicates that the sentiment in S1 is completely identical to the sentiment in S2, and 5 indicates that the sentiment has been completely transferred to the target sentiment in S2. |
| | S1: so he can charge a bloody fortune for them.<br>S2: so he can charge a fair amount of money for them. |
| | Sentiment transfer accuracy rating (on a scale of 1 to 5) = |

Table 9: A few-shot prompt for Sentiment Transfer Accuracy evaluation in Sentiment Transfer in English. It contains task definition, instruction, and input (see Section 3.2).

| Prompt | |
|---|---|
| | Sentiment transfer task: transfer the sentiment of a sentence (from positive to negative or negative to positive) while keeping the rest of the content unchanged. |
| | Please rate the content preservation between the following English source sentence S1 and the sentiment-transferred sentence S2 for the negative to positive sentiment transfer task on a scale of 1 to 5, where 1 indicates very low content preservation and 5 indicates very high content preservation. To determine the content preservation between these two sentences, consider only the information conveyed by the sentences and ignore any differences in sentiment due to the negative to positive sentiment transfer. |
| | S1: so he can charge a bloody fortune for them. |
| | S2: so he can charge a fair amount of money for them. |
| | Content Preservation rating (on a scale of 1 to 5) = |

Table 10: A few-shot prompt for Content Preservation evaluation in Sentiment Transfer in English. It contains task definition, instruction, and input (see Section 3.2).

| Prompt | |
|---|---|
| | Please rate the fluency of the following English sentence S on a scale of 1 to 5, where 1 represents poor fluency, and 5 represents excellent fluency. |
| | S: so he can charge a fair amount of money for them. |
| | Fluency rating (on a scale of 1 to 5) = |

Table 11: A few-shot prompt for Fluency evaluation in Sentiment Transfer in English. It contains instruction, and input (see Section 3.2).

## C  Pre-trained LLMs: Variants and Usage

This section describes the pre-trained Large Language Models (LLMs) used in our experiments. We detail their size variants (see Table 12) and specify the purposes for which they were used: zero-shot, few-shot, or fine-tuning (see Table 13).

## D  Human Evaluation Procedure

To evaluate the performance of our Text Sentiment Transfer models, we conducted a human evaluation focused on three critical aspects: *Style Transfer Accuracy*, *Content Preservation*, and *Fluency*. Below, we provide detailed definitions for each aspect and describe the questions used to guide the evaluation.

### D.1  Style Transfer Accuracy

**Definition:** Style Transfer Accuracy refers to how accurately the style of the original sentence has been transformed into the target sentiment. For instance, if a sentence originally expresses a negative sentiment, this metric evaluates whether it has been accurately converted to a positive sentiment, and vice versa.

**Evaluation Question:**

- *How accurately has the sentiment of the original sentence been transferred to the target sentiment?*

**Scoring:**

- **1**: No sentiment change; the original sentiment is entirely preserved.
- **2**: Minimal sentiment change; only slight evidence of sentiment transfer.
- **3**: Partial sentiment change; some aspects of the target sentiment are present, but the original sentiment still dominates.
- **4**: Considerable sentiment change; the target sentiment is clearly present, though traces of the original sentiment may remain.
- **5**: Complete sentiment change; the original sentiment has been entirely replaced by the target sentiment.

### D.2  Content Preservation

**Definition:** Content Preservation measures how well the style-independent meaning and core information of the original sentence are preserved after sentiment transfer.

**Evaluation Question:**

- *To what extent has the style-independent content and meaning of the original sentence been preserved after the sentiment transfer?*

| Model | Size Variants |
|---|---|
| BLOOM (BigScience Workshop, 2023) | 560M, 1B, 3B, and 7B |
| BLOOMz (Muennighoff et al., 2023) | 560M, 1B, 3B, and 7B |
| ChatGLM (Du et al., 2022) | 6B |
| ChatGLM2 (Du et al., 2022) | 6B |
| Falcon (Penedo et al., 2023; Almazrouei et al., 2023) | 7B |
| Llama (Touvron et al., 2023a) | 7B, 13B, and 30B |
| Llama-2 (Touvron et al., 2023b) | 7B, and 13B |
| Llama-2-Chat (Touvron et al., 2023b) | 7B, and 13B |
| Llama-3 (AI@Meta, 2024) | 8B |
| Llama-3-Instruct (AI@Meta, 2024) | 8B |
| Mistral-Instruct (Jiang et al., 2023) | 7B |
| OPT (Zhang et al., 2022) | 1.3B, 2.7B, 6.7B, 13B, and 30B |
| Zephyr (Tunstall et al., 2023) | 7B |

Table 12: List of open pre-trained LLMs used in our experiments, including their size variants.

| LLMs | Zero-shot | Few-shot | Finetuning |
|---|---|---|---|
| BLOOM-560M | ✓ | ✓ | ✓ |
| BLOOM-1B | ✓ | ✓ | ✓ |
| BLOOM-3B | ✓ | ✓ | ✓ |
| BLOOM-7B | ✓ | ✓ | ✓ |
| BLOOMz-560M | ✓ | ✓ | ✓ |
| BLOOMz-1B | ✓ | ✓ | ✓ |
| BLOOMz-3B | ✓ | ✓ | ✓ |
| BLOOMz-7B | ✓ | ✓ | ✓ |
| Falcon-7B | ✓ | ✓ | ✓ |
| ChatGLM-6B | ✓ | ✓ | ✗ |
| ChatGLM2-6B | ✓ | ✓ | ✓ |
| GPT-3.5 | ✓ | ✓ | ✗ |
| Llama-7B | ✓ | ✓ | ✓ |
| Llama-13B | ✓ | ✓ | ✓ |
| Llama-30B | ✓ | ✓ | ✗ |
| Llama-2-7B | ✓ | ✓ | ✓ |
| Llama-2-13B | ✓ | ✓ | ✓ |
| Llama-2-Chat-7B | ✓ | ✓ | ✗ |
| Llama-2-Chat-13B | ✓ | ✓ | ✗ |
| Llama-3-8B | ✓ | ✓ | ✓ |
| Llama-3-8B-Instruct | ✓ | ✓ | ✗ |
| Mistral-7B-Instruct | ✓ | ✓ | ✗ |
| OPT-1.7B | ✓ | ✓ | ✓ |
| OPT-2.7B | ✓ | ✓ | ✓ |
| OPT-6.7B | ✓ | ✓ | ✓ |
| OPT-13B | ✓ | ✓ | ✓ |
| OPT-30B | ✓ | ✓ | ✗ |
| Zephyr-7B | ✓ | ✗ | ✗ |

Table 13: Details of LLMs used for zero-shot, few-shot, or fine-tune scenarios. The model variant, including size and type (base/instructions/chat), is specified in the model name.

**Scoring:**

- **1**: Content is completely altered; the original meaning is lost.
- **2**: Major content changes; significant parts of the original meaning are altered or missing.
- **3**: Moderate content preservation; the general idea is retained, but with some noticeable changes.
- **4**: Good content preservation; most of the original meaning is intact with only minor alterations.
- **5**: Complete content preservation; the original meaning is fully retained.

### D.3 Fluency

**Definition:** Fluency assesses the grammatical correctness, naturalness, and overall readability of the sentence after the sentiment transfer. A fluent sentence should flow naturally and be free of awkward constructions or errors.

**Evaluation Question:**

- *How fluent and natural does the sentence sound after the sentiment transfer?*

**Scoring:**

- **1**: Not fluent at all; the sentence is grammatically incorrect and difficult to understand.
- **2**: Limited fluency; the sentence contains multiple errors and reads awkwardly.
- **3**: Moderate fluency; the sentence is somewhat understandable but has noticeable issues.
- **4**: Good fluency; the sentence is mostly clear with only minor issues.
- **5**: Complete fluency; the sentence is grammatically correct, natural, and easy to read.

### D.4 Evaluation Process

Evaluators are asked to rate each of these aspects on a 5-point Likert scale for a random sample of 50 sentences from the test set, equally split between positive-to-negative and negative-to-positive sentiment transfer tasks.

## E Additional Insights from Evaluation Results

In this section, we present a variety of graphs and charts to provide further insights into the automatic evaluation results, in addition to the analysis in Section 4. These visualizations are developed from Table 6. Additionally, we explore the correlation between GPT-4-based evaluations and human evaluations, as illustrated in Figure 1.

## F Sample Outputs from Top-Performing Models

In this section, we present a selection of sample outputs for the Detoxification task (Table 14). These outputs are generated from some of the best-performing models, as discussed in Section 4.2.

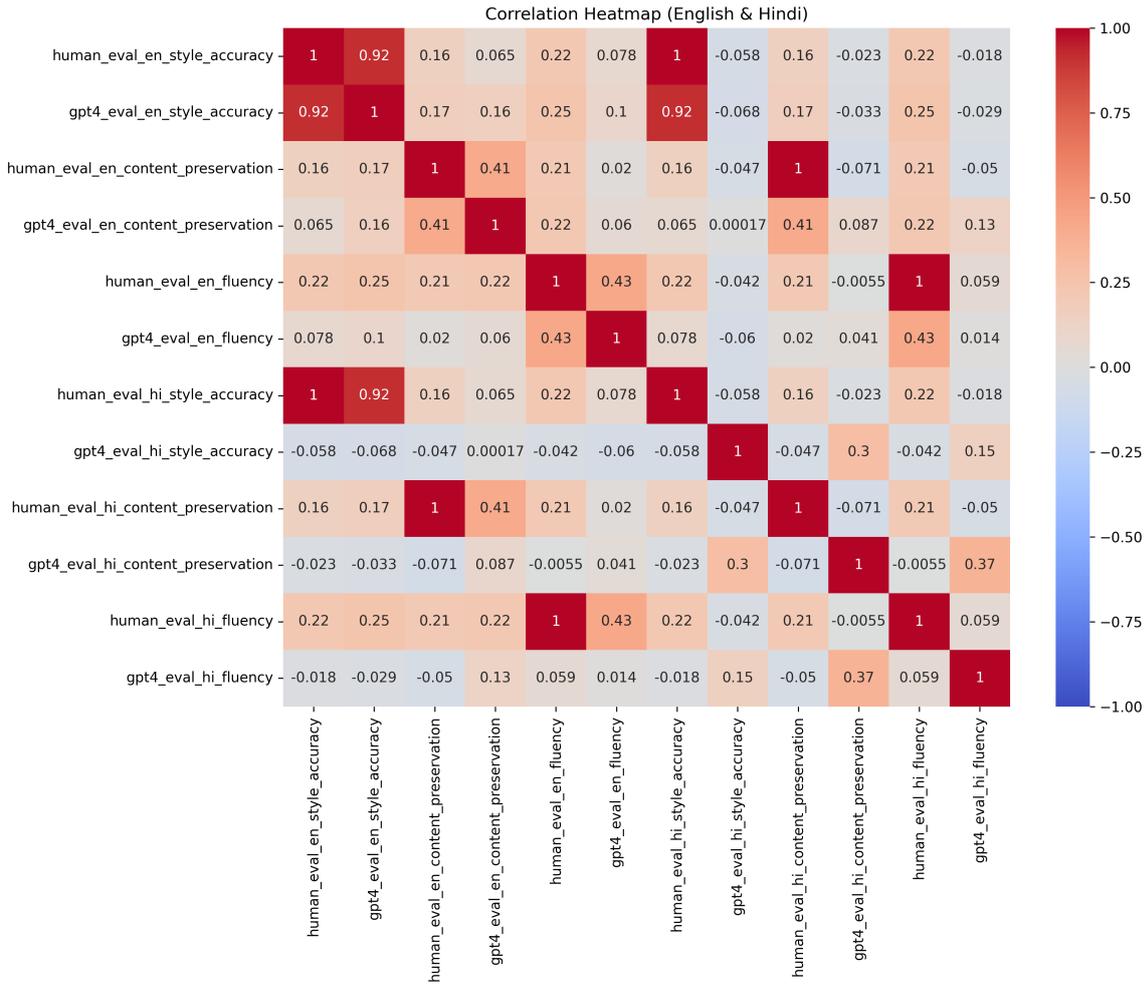

Figure 1: Correlation between GPT-4-based and Human Evaluation for sentiment transfer task in English and Hindi (see Section 4.2).

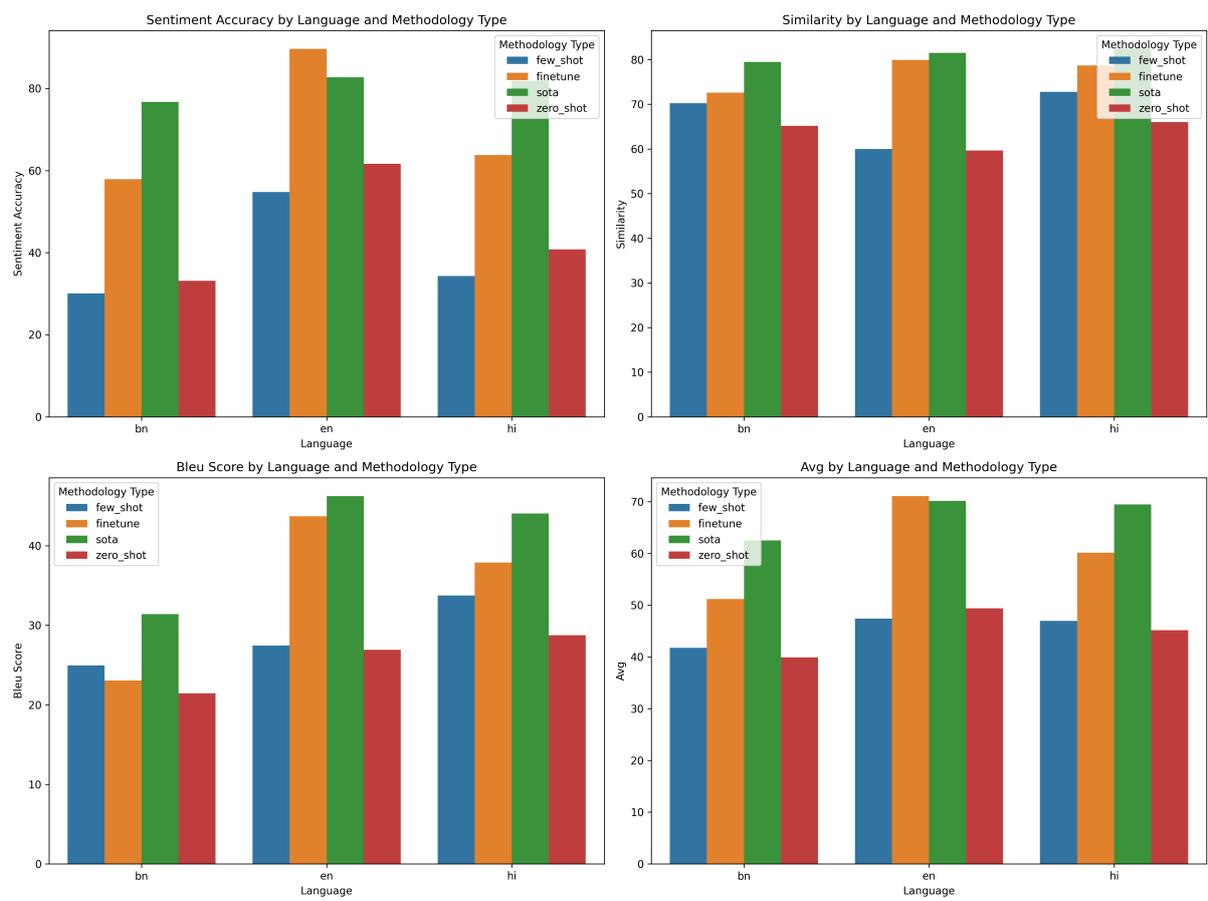

Figure 2: Comparison of various methodologies (zero-shot, few-shot, fine-tuning, and SOTA) by language for the Sentiment Transfer task across all models used (see Table 6).

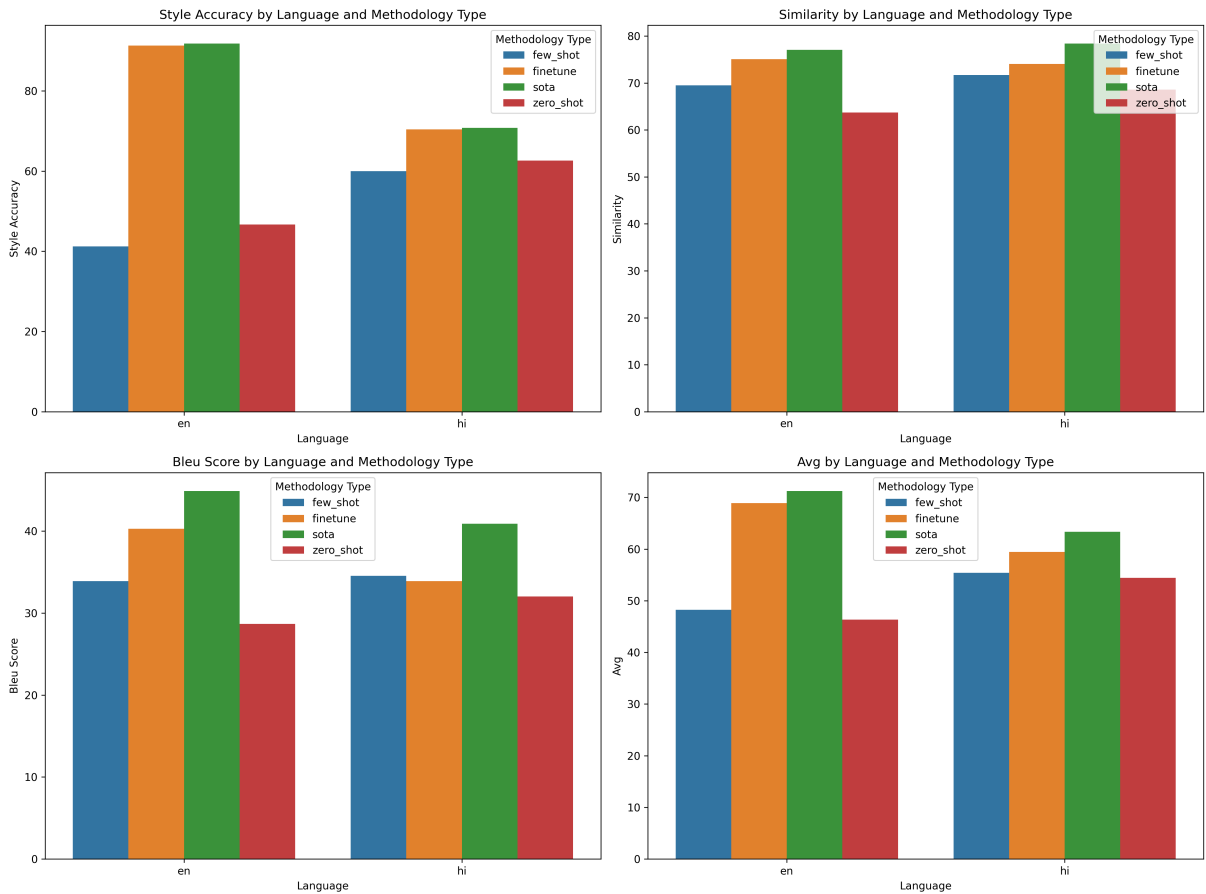

Figure 3: Comparison of various methodologies (zero-shot, few-shot, fine-tuning, and SOTA) by language for the Detoxification task across all models used (see Table 6).

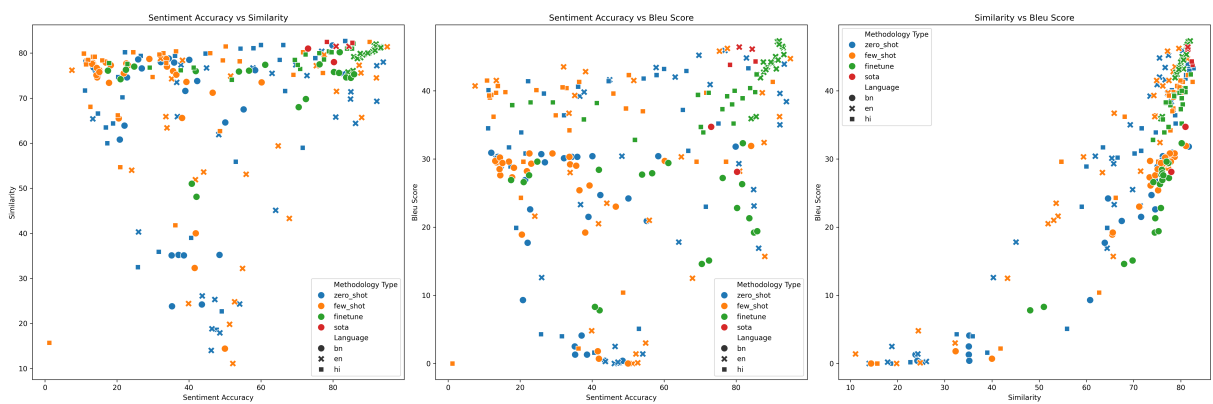

Figure 4: Relationships between (i) Sentiment Accuracy and BLEU Score, (ii) Sentiment Accuracy and Content Similarity, and (iii) BLEU Score and Content Similarity across zero-shot, few-shot, fine-tuning, and SOTA methodologies, spanning all languages in Sentiment Transfer task (see Table 6).

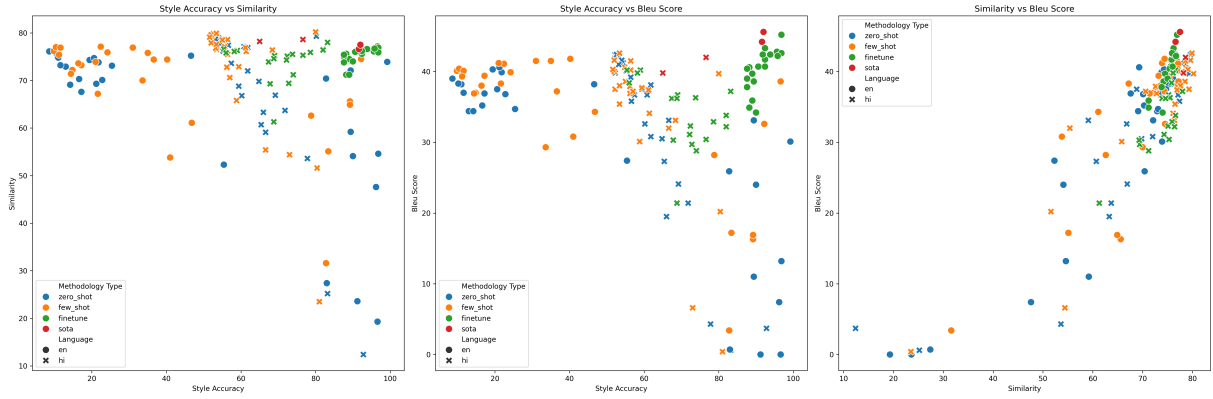

Figure 5: Relationships between (i) Sentiment Accuracy and BLEU Score, (ii) Sentiment Accuracy and Content Similarity, and (iii) BLEU Score and Content Similarity across zero-shot, few-shot, fine-tuning, and SOTA methodologies, spanning all languages in Detoxification task (see Table 6).

| Models | Toxic → Clean | Toxic → Clean |
|---|---|---|
| Reference | en: so shut the hell up until you know what it is your talking about . → So keep quiet untill you know what you are talking about. <br> hi: इसलिए अपना मूँह बंद रख जब तक तू नहीं जानता कि यह तू क्या बात कर रहा है। → इसलिए चुप रहो जब तक आप जानते नहीं कि आप किस बारे में बात कर रहे हैं। | en: p. s im a fucking cunt rag and enjoy the taste of a nice warm pile of shit every morning . → i enjoy the tasty nice warm pile every morning <br> hi: नोट करें: मैं एक नीच दर्जे की रंडी हूँ और हर रोज़ सुबह टट्टी के नाश्ते के मज़े लेती हूँ। → मैं हर रोज़ सुबह नाश्ते का मज़ा लेती हूँ। |
| GPT-3.5-ZS | en: Please refrain from speaking until you have a clear understanding of the topic. <br> hi: इसलिए अपना मूँह बंद रख जब तक तू नहीं जानता कि यह तू क्या बात कर रहा है। | en: p. s I am a person who enjoys the taste of a nice warm breakfast every morning. <br> hi: नोट करें: मैं एक नीच दर्जे की और हर रोज़ सुबह के नाश्ते के मज़े लेती हूँ। |
| Mistral-7B-Instruct-ZS | en: "So, shut up until you know what you're talking about." <br> hi: इसलिए अपना मूँह बंद रख जब तक तू नहीं जानता कि यह तू क्या बात कर रहा है। | en: p. s im a cunt rag and enjoy the taste of a nice warm pile of shit every morning . <br> hi: नोट करें: मैं एक नीच दर्जे की रंडी हूं और हर रोज़ सुबह टट्टी के नाश्ते के मज़े लेती हूं। |
| GPT-3.5-FS | en: so be quiet until you know what you're talking about. <br> hi: इसलिए अपना मूँह बंद रख जब तक तू नहीं जानता कि यह तू क्या कह रहा है। | en: P.S. I enjoy the taste of a nice warm pile of shit every morning. <br> hi: नोट करें: मैं एक नीच दर्जे की और हर रोज़ सुबह टट्टी के नाश्ते के मज़े लेती हूँ। |
| Mistral-7B-Instruct-FS | en: so shut up until you know what you're talking about. <br> hi: इसलिए अपना मूँह बंद रख जब तक तू नहीं जानता कि यह तू क्या बात कर रहा है। | en: p. s. i am a cunt rag and enjoy the taste of a nice warm pile of shit every morning. <br> hi: नोट करें: मैं एक नीच दर्जे की रंडी हूं और हर रोज़ सुबह टट्टी के नाश्ते के मज़े लेती हूं। |
| Llama-3-8B-FT | en: So please don't speak until you know what you are talking about. <br> hi: इसलिए अपना मुँह बंद रख जब तक आप यह नहीं जानते कि आप क्या बात कर रहे हैं। | en: p.s i am a bad person and enjoy the taste of a nice warm pile of shit every morning <br> hi: नोट करें: मैं एक नीच दर्जे की रंडी हूँ और हर रोज़ सुबह टट्टी के नाश्ते के मज़े लेती हूँ। |
| SOTA (KT) | en: so wait until you know what it is your talking about <br> hi: इसलिए अपना मूँह बंद रख जब तक तक तू नहीं जानता कि यह तू क्या बात कर रहा है | en: p. s im a bad guy and enjoy the taste of a nice warm pile of things every morning. <br> hi: मैं एक नीच दर्जे की रंडी हूँ और हर रोज़ सुबह टट्टी के नाश्ते के मज़ा लेती हूँ |

Table 14: Sample outputs from a few selected top-performing models (see Section 4.2) for the Text Detoxification task in English and Hindi are provided. Content warning: This table contains examples that are toxic, and/or offensive, and/or sexist in nature.